\documentclass[11pt]{article}

\usepackage{acl}

\usepackage{times}
\usepackage{latexsym}
\usepackage[T1]{fontenc}
\usepackage[utf8]{inputenc}
\usepackage{microtype}
\usepackage{courier}
\usepackage{graphicx}
\usepackage{booktabs}
\usepackage{amsmath}
\usepackage{amssymb}
\usepackage{tikz}
\usetikzlibrary{shapes.geometric, arrows.meta, positioning, fit, backgrounds, decorations.pathreplacing}

\definecolor{cascadeblue}{RGB}{0,106,166}
\hypersetup{colorlinks=true, citecolor=cascadeblue, linkcolor=cascadeblue, urlcolor=cascadeblue}
\newcommand{\kaonsicon}{\href{https://www.kaons.com/}{\textcolor{cascadeblue}{\raisebox{0.01ex}{\normalsize$\boldsymbol{K}^{\!*}$}}}}
\newcommand{\epochicon}{\href{https://www.epochlearn.com/}{\textcolor{cascadeblue}{\raisebox{0.02ex}{\normalsize$\boldsymbol{\phi}_{\!\mathbf{0}}$}}}}

\title{CascadeMind at SemEval-2026 Task 4: A Hybrid Neuro-Symbolic Cascade \\for Narrative Similarity}

\author{
\begin{tabular}{@{}c@{\hspace{1.55cm}}c@{}}
\textbf{Sebastien Kawada$^{\dagger,\ddagger}$} & \textbf{Dylan Holyoak$^{\ddagger}$} \\[0.45em]
{\normalfont $^{\dagger}$\,Kaons\,\kaonsicon} & {\normalfont $^{\ddagger}$\,Epoch Learn\,\epochicon} \\[0.35em]
{\normalfont\texttt{sebastien@kaons.com}} & {\normalfont\texttt{dylan@epochlearn.com}}
\end{tabular}}

\begin{document}
\maketitle

\begin{abstract}
Across self-consistency samples from an LLM, vote agreement tracks instance difficulty: on SemEval-2026 Task 4 (Narrative Story Similarity), supermajority cases ($\geq 7/8$ votes) resolve at 85\% accuracy, split votes at 67\%, and perfect ties at 61\%, a monotone gradient that holds across the development set. We exploit this in CascadeMind, which routes eight Gemini 2.5 Flash votes by consensus, escalates split votes to additional sampling rounds, and falls through to a symbolic ensemble of theory-inspired narrative signals only on perfect ties (5\% of cases). The system reached 72.75\% on Track A test, placing \textbf{10th of 44 teams}. Ablations show that the symbolic component contributes negligibly end-to-end and that nearly all gains come from confidence-aware routing. The takeaway is methodological: for narrative similarity, calibrating when to spend more compute on a hard instance matters more than adding auxiliary representations to reason about it. Code is available at \url{https://github.com/chreia/CascadeMind-ACL}.
\end{abstract}

\section{Introduction}

Comparing two stories for similarity goes beyond surface wording because it requires judging shared theme, sequence of events, and outcome \citep{piper2021narrative,tversky1977features}. SemEval-2026 Task 4 frames this as comparative judgment. Given an anchor and two candidate stories, systems decide which candidate is more similar to the anchor along three axes: abstract theme, course of action, and outcomes \citep{hatzel-etal-2026-semeval}.

LLMs handle most cases well but degrade on ambiguous comparisons \citep{keluskar2024llms}, making it important to know when to trust a single answer. CascadeMind treats vote agreement across self-consistency samples as a confidence signal and routes accordingly. Confident cases resolve fast, uncertain cases escalate, and perfect ties fall back to a symbolic ensemble. The system (1) samples eight self-consistency votes \citep{wang2022self} and treats vote distribution as an uncertainty signal, (2) commits when at least 7 of 8 votes agree, (3) escalates split votes to 32 votes total, and (4) falls back to a symbolic ensemble grounded in narrative theory only on perfect ties.

Our contributions are:
\begin{itemize}
    \item A cascade in which LLM vote agreement determines decision pathway, with pathway-level accuracy monotone in vote consensus (85\% / 67\% / 61\%)
    \item A symbolic ensemble of five theory-inspired similarity signals (lexical, story-grammar, semantic embedding, tension curve, event chain) used as a fallback only on perfect ties
    \item Ablations isolating the source of the gain, showing that confidence-aware routing accounts for nearly all of it while the symbolic fallback contributes negligibly because only 5\% of cases reach it
\end{itemize}

\section{Related Work}

\paragraph{Self-Consistency Decoding}
Self-consistency belongs to the ensemble tradition, aggregating multiple hypotheses to improve a prediction \citep{hansen2002neural}. \citet{wang2022self} introduced self-consistency as a decoding strategy that samples multiple reasoning paths and selects the most consistent answer through majority voting, with diverse reasoning traces yielding higher accuracy than single samples. Subsequent work treats sample consistency as a black-box uncertainty signal \citep{xiong2024uncertainty}. Extensions cover medical question answering \citep{maharjan2024openmedlm} and code generation \citep{huang2023enhancing}. Mirror-consistency addresses overconfidence in minority responses \citep{huang2024mirror}.

\paragraph{Narrative Representation Learning}
Computational approaches to narrative understanding draw on structural narrative theory as well as neural embedding representations. \citet{hatzel2024story} trained story embeddings on reformulations of the same story, making their setup the closest existing work on fictional-narrative similarity. The CoRRPUS framework \citep{dong2023corrpus} showed that structured code-based representations can improve story understanding in a neurosymbolic setting.

\paragraph{Ensemble Methods and Selective Prediction}
\citet{dietterich2000ensemble} characterized ensembles as weighted voters over component classifiers. Selective prediction allows models to abstain when uncertain, trading coverage for accuracy \citep{geifman2017selective}. Our cascade borrows the abstention trigger from selective prediction but always returns a prediction. Uncertainty selects pathway, not abstention.

\paragraph{Narrative Theory}
Our symbolic component draws on classical narrative theory. Propp's morphology of the folktale \citep{propp1968morphology} identified recurring structural elements across stories. Freytag's pyramid models narrative tension through exposition, rising action, climax, falling action, and resolution \citep{freytag1894technik}. Todorov's narrative grammar \citep{todorov1969grammaire} analyzes structured transformations in narrative. We operationalize these theories as computable similarity signals.

\section{System Architecture}
CascadeMind uses a direct comparative prompt asking which candidate story (A or B) is more similar to the anchor on abstract theme, course of action, and outcomes. The model returns JSON with a single \texttt{decision} field. No chain-of-thought rationale is requested or used.

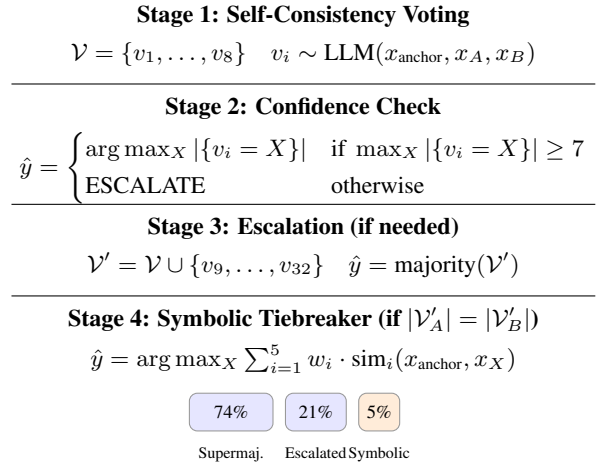
\begin{figure}[!htb]
\centering
\small
\setlength{\tabcolsep}{3pt}
\begin{tabular}{c}
\textbf{Stage 1: Self-Consistency Voting} \\[4pt]
$\mathcal{V} = \{v_1, \ldots, v_8\} \quad v_i \sim \text{LLM}(x_{\text{anchor}}, x_A, x_B)$ \\[6pt]
\hline \\[-6pt]
\textbf{Stage 2: Confidence Check} \\[6pt]
{\scriptsize$\hat{y} = \left\{
\begin{array}{@{}ll@{}}
\operatorname*{argmax}_{X} |\{v_i = X\}| & \text{if } \max_X |\{v_i = X\}| \geq 7 \\[2pt]
\textsc{Escalate} & \text{otherwise}
\end{array}
\right.$} \\[12pt]
\hline \\[-6pt]
\textbf{Stage 3: Escalation (if needed)} \\[4pt]
$\mathcal{V}' = \mathcal{V} \cup \{v_9, \ldots, v_{32}\} \quad \hat{y} = \text{majority}(\mathcal{V}')$ \\[6pt]
\hline \\[-6pt]
\textbf{Stage 4: Symbolic Tiebreaker (if $|\mathcal{V}'_A| = |\mathcal{V}'_B|$)} \\[4pt]
$\hat{y} = \arg\max_X \sum_{i=1}^{5} w_i \cdot \text{sim}_i(x_{\text{anchor}}, x_X)$
\end{tabular}

\vspace{6pt}
\begin{tikzpicture}[
    node distance=0.4cm,
    box/.style={rectangle, rounded corners=3pt, minimum height=0.5cm, draw=black!40, fill=#1, font=\scriptsize, text centered},
    arrow/.style={-{Stealth[scale=0.6]}, draw=black!40}
]
\node (s1) [box=blue!10, minimum width=1.15cm] {74\%};
\node (s2) [box=blue!10, minimum width=1.05cm, right=0.32cm of s1] {21\%};
\node (s3) [box=orange!15, minimum width=0.8cm, right=0.32cm of s2] {5\%};
\node[font=\tiny, below=0.1cm of s1] {Supermaj.};
\node[font=\tiny, below=0.1cm of s2] {Escalated};
\node[font=\tiny, below=0.1cm of s3] {Symbolic};
\end{tikzpicture}
\caption{Cascade decision process. Most cases (74\%) resolve via supermajority, 21\% require escalation, and 5\% invoke the symbolic ensemble.}
\label{fig:architecture}
\end{figure}

The four-stage pipeline lets neural voting handle most cases while a symbolic fallback engages only when voting fails, as shown in Figure~\ref{fig:architecture}.

\subsection{Neural Self-Consistency Voting}

We use Gemini 2.5 Flash \citep{google-gemini-2-5-flash-2025} with \texttt{candidateCount=8} to return eight responses per call through the Gemini API \citep{google-generative-ai-api-2025}. Each response is a single A/B vote.

\paragraph{Decision Logic}
Let $V = \{v_1, \ldots, v_8\}$ be the set of votes where $v_i \in \{A, B\}$. We define the vote count $c_X = |\{v_i : v_i = X\}|$ for $X \in \{A, B\}$.

\textbf{Supermajority:} If $\max(c_A, c_B) \geq 7$, we return the majority decision immediately.

\textbf{Escalation:} On splits (4-4, 5-3, or 6-2), we issue three additional \texttt{candidateCount=8} calls for 32 votes total and take the majority. A perfect 16-16 tie triggers the symbolic fallback.

\subsection{Multi-Scale Narrative Analysis Ensemble}

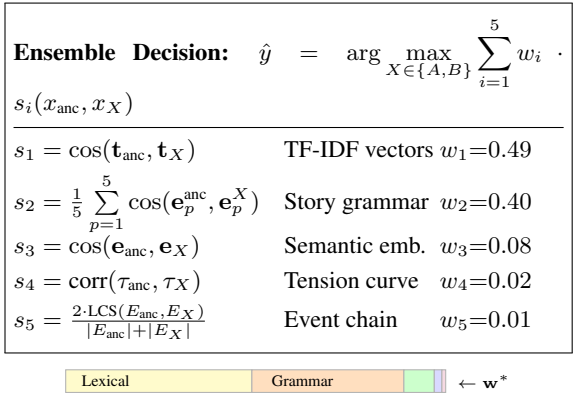
\begin{figure}[!htb]
\centering
\small
\fbox{\parbox{0.94\columnwidth}{
\vspace{2pt}
\textbf{Ensemble Decision:}
\vspace{-6pt}
\[
\hat{y}=\operatorname*{argmax}_{X\in\{A,B\}}\sum_{i=1}^{5}w_i\,s_i(x_{\mathrm{anc}},x_X)
\]
\vspace{-8pt}
\vspace{4pt}
\hrule
\vspace{4pt}
\begin{tabular}{@{}r@{\;}l@{\quad}l@{}}
$s_1$ & $= \cos(\mathbf{t}_{\text{anc}}, \mathbf{t}_X)$ & {\footnotesize TF-IDF vectors \hfill $w_1{=}0.49$} \\[3pt]
$s_2$ & $= \frac{1}{5}\sum\limits_{p=1}^{5} \cos(\mathbf{e}_{p}^{\text{anc}}, \mathbf{e}_{p}^{X})$ & {\footnotesize Story grammar \hfill $w_2{=}0.40$} \\[5pt]
$s_3$ & $= \cos(\mathbf{e}_{\text{anc}}, \mathbf{e}_X)$ & {\footnotesize Semantic emb. \hfill $w_3{=}0.08$} \\[3pt]
$s_4$ & $= \text{corr}(\tau_{\text{anc}}, \tau_X)$ & {\footnotesize Tension curve \hfill $w_4{=}0.02$} \\[3pt]
$s_5$ & $= \frac{2 \cdot \text{LCS}(E_{\text{anc}}, E_X)}{|E_{\text{anc}}| + |E_X|}$ & {\footnotesize Event chain \hfill $w_5{=}0.01$}
\end{tabular}
\vspace{2pt}
}}
\vspace{4pt}

\begin{tikzpicture}
\draw[fill=yellow!25, draw=black!30] (0,0) rectangle (2.45,0.3);
\draw[fill=orange!25, draw=black!30] (2.45,0) rectangle (4.45,0.3);
\draw[fill=green!20, draw=black!30] (4.45,0) rectangle (4.85,0.3);
\draw[fill=blue!15, draw=black!30] (4.85,0) rectangle (4.95,0.3);
\draw[fill=purple!15, draw=black!30] (4.95,0) rectangle (5,0.3);
\node[font=\tiny, anchor=west] at (0.1,0.15) {Lexical};
\node[font=\tiny, anchor=west] at (2.6,0.15) {Grammar};
\node[font=\tiny] at (5.5,0.15) {$\leftarrow \mathbf{w}^*$};
\end{tikzpicture}

\caption{Multi-Scale Narrative Ensemble with five weighted similarity signals ($\mathbf{w}^*$ optimized via differential evolution).}
\label{fig:ensemble}
\end{figure}

On perfect ties after escalation, the system falls back to a symbolic ensemble of five similarity signals at different levels of abstraction (Figure~\ref{fig:ensemble}). Let $s_i^A$ and $s_i^B$ denote the similarity scores between the anchor and stories A and B for signal $i$.

\paragraph{Signal 1: Lexical Similarity (TF-IDF)}
We TF-IDF-vectorize each story \citep{salton1988term} and compute cosine similarity between the anchor and each candidate:
\begin{equation}
s_{\text{lex}}^X = \cos(\text{tfidf}(\text{anchor}), \text{tfidf}(X))
\end{equation}
This captures surface lexical overlap, including shared characters and domain terminology.

\paragraph{Signal 2: Story Grammar Similarity}
As a heuristic inspired by story-grammar and narrative-phase models \citep{propp1968morphology,freytag1894technik,thorndyke1977storygrammar}, we segment each story into five narrative phases based on position: setting (first 20\%), conflict (20-40\%), rising action (40-60\%), climax (60-80\%), and resolution (80-100\%). We compute a sentence-transformer embedding for each aligned phase and average cosine similarities across the five phases:
\begin{equation}
s_{\text{grammar}}^X = \frac{1}{5}\sum_{p \in P} \cos(\text{enc}(a_p), \text{enc}(x_p))
\end{equation}
where $P$ = \{setting, conflict, rising, climax, resolution\}.

\paragraph{Signal 3: Semantic Similarity}
We encode each full story with all-MiniLM-L6-v2 \citep{reimers2019sentence,reimers2020allminilm} and compute cosine similarity:
\begin{equation}
s_{\text{sem}}^X = \cos(\text{enc}(\text{anchor}), \text{enc}(X))
\end{equation}
This complements the phase-aligned signal with whole-story semantics.

\paragraph{Signal 4: Narrative Tension Curve}
Inspired by Freytag's pyramid \citep{freytag1894technik} and computational work on sentiment-derived story arcs \citep{reagan2016emotionalarcs}, we use a heuristic positional tension proxy. For each sentence, we compute a tension score as the sum of sentiment intensity (absolute polarity) and subjectivity using TextBlob \citep{loria2018textblob}. Per-sentence tension is linearly interpolated to a 10-point curve, and we then compute Pearson correlation between curves:
\begin{equation}
s_{\text{tension}}^X = \text{corr}(T_{\text{anchor}}, T_X)
\end{equation}
This procedure captures similarity in emotional dynamics and pacing.

\paragraph{Signal 5: Event Chain Similarity}
We extract action verbs using spaCy's part-of-speech tagger \citep{spacy-2025}, filtering to verbs matching a list of 47 narrative action words inspired by event-chain modeling and Propp's character functions \citep{chambers2008narrativechains,propp1968morphology} (e.g., \emph{discover}, \emph{fight}, \emph{escape}, \emph{transform}). These form event sequences $E_{\text{anc}}$ and $E_X$ (ordered verb lemmas). We compute the longest common subsequence (LCS):
\begin{equation}
s_{\text{event}}^X = \frac{2 \cdot \text{LCS}(E_{\text{anc}}, E_X)}{|E_{\text{anc}}| + |E_X|}
\end{equation}
This procedure approximates plot-structure similarity using ordered action-verb overlap.

\paragraph{Weighted Ensemble}
The final score is a weighted sum:
\begin{equation}
\text{score}^X = \sum_{i=1}^{5} w_i \cdot s_i^X
\end{equation}
The ensemble returns A if $\text{score}^A > \text{score}^B$ and B otherwise, with exact ties defaulting to B in our implementation.

We fit $w_i$ via differential evolution \citep{storn1997differential} on the organizers' 1,900-triplet synthetic split \citep{narrative-similarity-data-2026}. This split is LLM-generated and is used only for calibrating symbolic weights. Development and test reporting in this paper use the task's human-labeled splits.

The optimization objective minimizes classification error:
\begin{equation}
\mathbf{w}^* = \arg\min_{\mathbf{w}} \sum_{j=1}^{N} \mathbf{1}\Big[\text{sign}\big(\textstyle\sum_{i} w_i \Delta s_i^j\big) \neq y_j\Big]
\end{equation}
where $\Delta s_i^j = s_i^A - s_i^B$ and $y_j \in \{-1, +1\}$ is the ground truth label. We use differential evolution due to its effectiveness on non-convex, non-differentiable objectives. 

The optimized weights are shown in Figure~\ref{fig:ensemble}. Lexical (49\%) and story-grammar (40\%) signals dominate, suggesting the synthetic training data discriminates on surface wording and phase structure rather than deeper semantics. The ensemble achieves 99.5\% accuracy on a held-out synthetic validation split (n=200).

\section{Experiments}

\subsection{Dataset}

The SemEval-2026 Task 4 dataset and evaluation protocol are described by the task organizers \citep{hatzel-etal-2026-semeval,narrative-similarity-data-2026}.\footnote{Shared-task data and generated submission artifacts should be retrieved from the task source unless redistribution permission is confirmed.} All story summaries and labels used in this work are in English. The task uses Wikipedia plot synopses from the English portion of Tell-Me-Again, filtered to short summaries (typically four to eight sentences). We use the 200-triplet Track A development split for analysis and the 400-triplet test split for the official submission. Each triplet is one anchor with two candidates and a label naming the closer candidate. The synthetic training set comprises 1,900 LLM-generated triplets and is used only for symbolic-weight calibration.

\subsection{Implementation Details}

We use Gemini 2.5 Flash via the Google AI API with \texttt{candidateCount=8} for multi-candidate voting. Temperature is 1.0 to encourage diversity, and other parameters use defaults. The symbolic ensemble module uses scikit-learn for TF-IDF vectorization \citep{pedregosa2011scikit}, sentence-transformers (all-MiniLM-L6-v2) for embeddings, and TextBlob for sentiment analysis.

\subsection{Results}

\subsubsection{Official Shared-Task Result}
Table~\ref{tab:official_results} reports our official result \citep{narrative-similarity-results-2026}. The task received 71 submissions from 46 teams across both tracks, and Track A had 44 ranked teams.

\begin{table}[t]
\centering
\small
\begin{tabular}{lccc}
\toprule
\textbf{System} & \textbf{Track} & \textbf{Test Accuracy} & \textbf{Rank} \\
\midrule
CascadeMind & A & 72.75\% & \textbf{10th/44} \\
\bottomrule
\end{tabular}
\caption{Official shared-task result for our final submission. Rank is among the 44 Track A teams.}
\label{tab:official_results}
\end{table}

\subsubsection{Post-Hoc Test-Set Analysis}
After release of Track A test labels \citep{narrative-similarity-dataset-2026}, we computed post-hoc test-set diagnostics for CascadeMind on all 400 examples. The official leaderboard score remains the submitted 72.75\% (291/400) in Table~\ref{tab:official_results}. The post-hoc diagnostic predictions score 73.0\% accuracy (292/400), a one-prediction difference.

\begin{table}[t]
\centering
\small
\begin{tabular}{lrr}
\toprule
\textbf{Gold Label} & \textbf{Pred A} & \textbf{Pred B} \\
\midrule
A closer ($n{=}208$) & 143 & 65 \\
B closer ($n{=}192$) & 43 & 149 \\
\bottomrule
\end{tabular}
\caption{Post-hoc confusion matrix on released Track A test labels (n=400).}
\label{tab:test_confusion}
\end{table}

Class-wise behavior is asymmetric. For the \emph{A-closer} class, precision/recall/F1 are 76.9\%/68.8\%/72.6\%. For \emph{B-closer}, they are 69.6\%/77.6\%/73.4\%. Macro-F1 is 73.0\% and balanced accuracy is 73.2\%. The model predicts B more often than A (53.5\% vs. 46.5\%), while the released label distribution is slightly A-leaning (52.0\% A, 48.0\% B), indicating a mild tendency to over-select B in uncertain comparisons.

\subsubsection{Development-Set Analysis}
Table~\ref{tab:dev_results} presents development diagnostics with different denominators. Baselines are computed on the full development split (n=200), while cascade routing diagnostics are computed on a separate subset (n=100). CascadeMind reaches 81.0\% on the cascade subset. Because these denominators differ, cross-block percentage-point differences are descriptive only. With 74\% resolved at supermajority and 26\% escalated to three additional calls, expected API usage is 1.78 calls per case.

All reported LLM decisions are direct A/B outputs. No chain-of-thought rationales are requested, exposed, or used.

\begin{table}[t]
\centering
\footnotesize
\begin{tabular}{@{}p{0.50\linewidth}cc@{}}
\toprule
\textbf{System} & \textbf{Accuracy} & \textbf{Calls/Case} \\
\midrule
Single Vote & 68.0\% & 1.0 \\
Self-Consistency (k=8) & 76.5\% & 1.0 \\
Majority (k=3 calls) & 78.0\% & 3.0 \\
\midrule
CascadeMind & 81.0\% & 1.78 avg \\
\quad + Symbolic Tiebreaker & 81.0\% & 1.78 avg \\
\bottomrule
\end{tabular}
\caption{Development-set system comparison (n=200 for baselines, n=100 for cascade experiments).}
\label{tab:dev_results}
\end{table}

\subsubsection{Performance by Decision Pathway}
Table~\ref{tab:ablation} breaks down accuracy by pathway. Supermajority cases (74\% of the development subset) reach 85\%, supporting vote consensus as a confidence signal. Escalated cases resolve at 67\% under majority, and a separate perfect-tie diagnostic set reaches 61\% after symbolic processing. Lower vote consensus tracks lower accuracy, consistent with multi-sample consistency as an uncertainty signal \citep{xiong2024uncertainty}.

\begin{table}[t]
\centering
\footnotesize
\begin{tabular}{@{}p{0.47\linewidth}cc@{}}
\toprule
\textbf{Pathway} & \textbf{Cases} & \textbf{Accuracy} \\
\midrule
Supermajority ($\geq$7/8) & 74/100 & 85\% \\
Escalated (majority) & 21/100 & 67\% \\
Escalated (symbolic tie) & 5/100 & 61\% (11/18) \\
\midrule
\textit{Total} & \textit{100/100} & \textit{81\%} \\
\bottomrule
\end{tabular}
\caption{Performance breakdown by decision pathway. Routing shares are measured on the cascade diagnostic subset (n=100). Symbolic-tie accuracy is measured on a separate perfect-tie diagnostic set (n=18).}
\label{tab:ablation}
\end{table}

\subsubsection{Symbolic Tiebreaker Analysis}

On the separate perfect-tie diagnostic set (n=18), the symbolic tiebreaker achieves 61.1\% accuracy (11/18), as shown in Table~\ref{tab:ablation}. Manual inspection shows recurring errors involving misleading lexical overlap, non-standard narrative structures that violate positional assumptions, and edge cases that require world knowledge beyond surface signals. Applied to all cases regardless of neural confidence, the symbolic ensemble drops to 53\% on the cascade diagnostic subset, making it suitable only as a high-uncertainty fallback.

\begin{table}[t]
\centering
\footnotesize
\begin{tabular}{@{}lcccc@{}}
\toprule
\textbf{Split} & \textbf{n} & \textbf{Acc.} & \textbf{F1} & \textbf{A/B} \\
\midrule
Dev & 200 & 57.0\% & 57.0 & 99/101 \\
Test & 400 & 60.5\% & 60.4 & 208/192 \\
\bottomrule
\end{tabular}
\caption{Post-hoc symbolic-only support check using the fixed paper weights and no LLM calls.}
\label{tab:symbolic_only}
\end{table}

Table~\ref{tab:symbolic_only} confirms above-chance behavior of the symbolic module as a standalone classifier, scoring 57.0\% on development (114/200) and 60.5\% on test (242/400). Prediction rates are balanced on both splits, but absolute accuracy stays below the neural cascade. Voting and escalation, not symbolic signals, drive performance.

\section{Discussion}

\paragraph{Test-Time Behavior}
Table~\ref{tab:official_results} summarizes official shared-task standing, and Table~\ref{tab:test_confusion} shows class-level behavior for the post-hoc Track A predictions. The dominant test-time error is predicting B when A is correct (65 cases), which is consistent with lower recall on the A-closer class than on the B-closer class.

\paragraph{Value of Cascade Design}
On the development set, the cascade's gains come from voting and escalation, not the symbolic tiebreaker. The supermajority rule identifies confident predictions cheaply (85\% on 74\% of cases), and escalation adds signal for borderline cases. LLM vote distribution serves as a useful uncertainty indicator in this setting because higher consensus correlates with higher accuracy, consistent with broader evidence on sample consistency as black-box LLM uncertainty \citep{xiong2024uncertainty}.

\paragraph{Domain Mismatch}
The symbolic ensemble achieves 99.5\% accuracy on a held-out synthetic validation split but only 61.1\% on the perfect-tie diagnostic set, and overall system performance drops from 81.0\% on the cascade diagnostic subset to 72.75\% on official test. This gap is consistent with distribution mismatch, overfitting to synthetic calibration data, or greater difficulty in the final shared-task setting.

\paragraph{Signal Complementarity}
In this symbolic ensemble trained on synthetic data, weights concentrate on lexical (49\%) and story-grammar (40\%) signals over semantic embeddings (8\%). The 8\% semantic weight is small but nonzero, capturing information not covered by the lexical and structural signals.

\paragraph{Measure Limitations}
The 1\% weight on event-chain similarity likely reflects sparsity from exact verb matching against a 47-word list. Richer event representations---fuzzy matching, full narrative event chains \citep{chambers2008narrativechains}, or semantic role labeling---are a natural next step.

\section{Conclusion}

We presented a hybrid neuro-symbolic cascade model for narrative story similarity that combines neural self-consistency-style voting with a Multi-Scale Narrative Analysis Ensemble. In official shared-task evaluation, CascadeMind reaches 72.75\% Track A test accuracy (Table~\ref{tab:official_results}). On development diagnostics (cascade subset, n=100), it reaches 81.0\%.

The main contributions are threefold. First, CascadeMind shows that LLM vote distribution can act as a useful uncertainty indicator, with supermajority predictions reaching 85\% accuracy on development. Second, ablations isolate where the gain comes from, showing that confidence-aware routing accounts for nearly all of it while the symbolic tiebreaker contributes negligibly because only 5\% of cases reach it. Third, the paper documents a development-to-test gap (81.0\% to 72.75\%) in shared-task conditions.

\section*{Limitations}

The system depends on a commercial API (Gemini 2.5 Flash) and stochastic decoding. While the routing policy and thresholds are fixed, exact vote distributions can vary across runs and model revisions, which limits strict reproducibility without pinned model snapshots \citep{pineau2021reproducibility,chen2024chatgptbehavior}.

The development-to-test gap is substantial. Accuracy is 81.0\% on the cascade diagnostic subset (n=100) versus 72.75\% (291/400) in official shared-task evaluation. Post-hoc Track A diagnostics score 73.0\% (292/400), a one-prediction difference from the submitted file.

The symbolic tiebreaker has narrow end-to-end impact because only 5\% of the cascade diagnostic subset reaches the final tie state. Most gains come from neural voting and escalation rather than symbolic resolution, so improvements to confidence calibration and escalation strategy are likely to matter more than additional symbolic feature engineering.

Post-hoc test analysis shows asymmetric class behavior. Recall is lower for A-closer cases (68.8\%) than for B-closer cases (77.6\%), and the model predicts B slightly more often than the label distribution warrants. Reducing this decision bias is a direct target for future versions.

\section*{Ethics Statement}

This work uses public story summaries and organizer-provided labels. We did not recruit, interact with, or collect data from human subjects. The LLM component may inherit biases and other risks present in its training data and commercial LLM deployment settings \citep{weidinger2022taxonomy}.

\section*{Acknowledgements}

We thank the SemEval-2026 Task 4 organizers for creating the benchmark and running the shared task.

\bibliography{references}

\end{document}